\pdfoutput=1

\documentclass[11pt]{article}

\usepackage{EACL2023}
\usepackage{tabularx}

\usepackage{times}
\usepackage{latexsym}
\usepackage{array}
\usepackage{makecell}
\usepackage{graphicx}
\usepackage[T1]{fontenc}

\usepackage[utf8]{inputenc}

\usepackage{microtype}

\usepackage{inconsolata}
\usepackage{amsmath}

\title{TR-EduVSum: A Turkish-Focused Dataset and Consensus Framework for Educational Video Summarization}

\author{
Figen E\u{g}in \\
Department of Computer Engineering \\
Izmir Katip Celebi University \\
Cigli, Izmir, T\"urkiye \\
\texttt{figenkaya@gmail.com}
\And
Aytu\u{g} Onan \\
Department of Computer Engineering \\
Izmir Institute of Technology \\
Urla, Izmir, T\"urkiye \\
\texttt{aytugonan@iyte.edu.tr}
}

\begin{document}
\maketitle
\begin{abstract}
This study presents a framework for generating the gold-standard summary fully automatically and reproducibly based on multiple human summaries of Turkish educational videos. Within the scope of the study, a new dataset called TR-EduVSum was created, encompassing 82 Turkish course videos in the field of ``Data Structures and Algorithms'' and containing a total of 3,281 independent human summaries. Inspired by existing pyramid-based evaluation approaches, the AutoMUP (Automatic Meaning Unit Pyramid) method is proposed, which extracts consensus-based content from multiple human summaries. AutoMUP clusters the meaning units extracted from human summaries using embedding, statistically models inter-participant agreement, and generates graded summaries based on consensus weight. In this framework, the gold summary corresponds to the highest-consensus AutoMUP configuration, constructed from the most frequently supported meaning units across human summaries. Experimental results show that AutoMUP summaries exhibit high semantic overlap with robust LLM (Large Language Model) summaries such as Flash~2.5 and GPT-5.1. Furthermore, ablation studies clearly demonstrate the decisive role of consensus weight and clustering in determining summary quality. The proposed approach can be generalized to other Turkic languages at low cost.
\end{abstract}

\section{Introduction}

The amount of video content shared online has reached enormous proportions. Video format has also become widespread in educational content, and numerous channels have emerged on platforms like YouTube, publishing educational videos in various fields. Learning a subject through video lessons has both advantages and disadvantages. Multimedia content facilitates a better understanding of the subject. However, watching the content can be time-consuming, and in some cases, the content may not actually convey the intended concepts \citep{herrington2025}. Providing summaries of these videos to the user allows for detailed information about the content without requiring the user to watch the video itself, thereby overcoming the problem of finding the actual content. Video summarization presents a challenging problem in summarization studies because it encompasses various elements, including audio and video. Furthermore, text summarization models may struggle with spoken language tasks \citep{lv2021}. Moreover, in educational videos, the summary of the transcription is often insufficient for understanding the content. To make video summarization more effective, multimodal video summarization studies that process different data types simultaneously are emerging \citep{huang2024,zhu2023}. These studies combine the power of computer vision and NLP models to enable the production of more meaningful summaries. Studies presenting a comparative evaluation of multimodal approaches reveal that components such as modality fusion, feature selection, and segment consistency directly affect the quality of summarization \citep{marevac2025}. Some studies evaluating automated summarization systems focus on accurately identifying the information to be summarized. One such approach, which involves evaluating meaning units and producing video summaries accordingly, is exemplified by methods such as Pyramid, LitePyramid, and ACU \citep{nenkova2004,shapira2019,liu2023}. In these methods, content units are extracted by human annotators and matched with the system summary. These units are weighted according to the frequency with which they appear in human summaries. Lite3Pyramid \citep{zhang2021lite3}, created with a similar logic, is one of the first models to automate Pyramid and LitePyramid. Semantic Role Labeling (SRL) units and Natural Language Inference (NLI), which checks whether a content unit is present in the system summary, are included. QAPyramid, another method inspired by the pyramid method, is a reference-based summary evaluation metric that breaks down the summary into question-and-answer sections to clarify content units \citep{zhang2024qapyramid}.

With the emergence of LLMs, a more flexible and human-like quality of summarization has been achieved \citep{jin2024}. However, this method is still under development. Although all these methods represent significant advancements in video summarization, they have considerable limitations in evaluating educational videos and generating gold summaries. Content unit-based evaluation approaches, such as Pyramid, LitePyramid, and ACU, offer high accuracy but are entirely dependent on human annotation. Extracting, grouping, and matching semantic units, such as SCU or ACU, with system summaries requires expertise and is quite costly to implement on large-scale datasets. Furthermore, the gold content units and significance weights produced by these methods are susceptible to annotator bias and inconsistencies in human interpretation. Lite3Pyramid reduces this cost because it is an automated system. However, it may not work reliably with long, structurally complex texts such as educational video transcripts. Therefore, even though the mechanism proposed by Lite3Pyramid is scalable, it cannot be generalized to domain-specific scenarios.

Approaches like LLM-Pyramid are powerful in capturing complex semantic relationships. However, structural errors in LLMs, such as hallucinations, inconsistencies, and model bias, reduce the reliability of these methods. Furthermore, the decision-making process is not transparent, resulting in outcomes that depend on the model version and prompt. This creates reproducibility problems and prevents LLMs from being used as a gold standard. In this context, the current literature lacks a gold standard framework that generates video summaries in a completely unsupervised and reproducible manner, is largely language-agnostic given the availability of suitable multilingual embeddings, statistically models human consensus, and is both low-cost and free from LLM bias. This gap is particularly evident for Turkish-specific video summarization datasets.

Turkish and other Turkic languages are more amenable to summarizing duplicate content in different ways due to their morphological structure and high variety of expressions, which makes modeling and evaluation difficult. Therefore, there is a continuing need for a fully automated and repeatable consensus-based summary generation framework that incorporates multiple human summaries, particularly for Turkish educational video summarization datasets. Based on this need, this study presents a dataset of Turkish educational videos created using human-generated summaries. Furthermore, we introduce an automated and scalable counterpart to human-intensive Pyramid paradigms for gold summary generation.

The contributions of this article are summarized as follows:
\begin{itemize}
    \item The TR-EduVSum dataset, containing multiple human summaries, is introduced for Turkish educational video summarization.
    \item This study is among the first to explore semantic unit--based methods for Turkish educational video summarization datasets, to our knowledge.
    \item AutoMUP, a new framework for generating gold summaries from multiple human summaries, is proposed.
\end{itemize}

The proposed framework generates gold summaries reflecting different confidence levels by stratifying content units according to the degree of consensus across multiple human summaries. Within the scope of the study, a dataset of 82 Turkish educational videos was created, with each video independently summarized by multiple participants, resulting in 36 to 53 human-written summaries per video. The dataset provides a rich basis for both gold summary generation and summary evaluation. Although the study was conducted in Turkish, its largely language-agnostic structure allows it to be replicated in other Turkic languages at low cost.

\section{Related Work}

\subsection{Educational Video Summarization}

Summarizing educational videos is beneficial for students, as it ensures they have access to the correct content and allows them to review and recall fundamental concepts. While many tools exist for summarizing, summarizing video lectures is challenging due to their complex structures and lengthy formats \citep{xie2025}. The fact that video lectures often contain spoken language is another factor that makes summarizing difficult. The summary should encompass not only the transcript but also the visually presented materials and practical exercises. This deficiency is being addressed through a combination of multimodal approaches.

One such approach, MF2Summ, is a multimodal model that uses both visual and auditory information in the video summarization task \citep{wang2024}. This model extracts visual features using GoogLeNet and audio features using SoundNet, and then combines these two modalities using a cross-modal Transformer. REFLECTSUMM was developed for summarizing student lecture reflections \citep{zhong2024}. It provides a strong benchmark for educational technologies, but it is text-focused and does not include video, spoken language, or multimodal content. A similar study, VT-SSum, is a large-scale benchmark aiming at segmentation and extractive summarization for transcripts of 9,616 educational videos \citep{lv2021}. In this study, slide content is assumed to be the gold ratio. This, combined with automatic speech errors, can negatively impact the quality of the summaries. Overall, studies have focused on the English language, and research on lecture videos is limited.

\subsection{Turkic Summarization and Evaluation}

Studies on evaluating video summaries vary. SEval-Ex offers a framework that provides both high accuracy and explainability by reducing summary evaluation to the atomic level \citep{herserant2025}. VSUMM proposes a new evaluation method based on human-generated summaries, in which automated summaries are compared with human summaries and error rates are examined \citep{avila2011}.

However, studies on video summarization in Turkic languages are quite limited. While no study has been found that focuses on creating a Turkish video summarization dataset, a Turkish video captioning dataset was created by translating from the original English dataset. MSVD-Turkish includes descriptions of short video clips and reports features reflecting the agglutinative structure of Turkish \citep{citamak2021}. Another summarization study in Turkish was conducted by Erdağı and Tunalı \citep{erdagi2024}. In this study, feature-based sentence ordering methods are compared for Turkish news text summarization. The results show that a hybrid approach yields the best performance and that the methods can produce results close to robust models such as BERTSum.

Fikri et al.\ \citep{fikri2021} stated that the ROUGE metric is not suitable for evaluating abstractive summarization systems because it is based on lexical overlap between summaries produced using the gold standard. The authors translated the English STSb dataset into Turkish and presented the first semantic textual similarity dataset for the Turkish language. Deep reinforcement learning-based approaches for Turkish abstractive summarization have also been conducted \citep{fikri2024}. New evaluation criteria based on semantic similarity calculated with BERTurk have been presented, and it has been shown that these criteria provide a higher correlation with human evaluations. Furthermore, it has been shown that a hybrid model trained using these semantic similarity scores as a reward function produces more natural and readable summaries.

These studies on summarization methods and evaluation criteria offer significant advancements for Turkish, but clearly highlight the lack of resources in the field of Turkish video summarization. This deficiency creates a significant gap in the training and evaluation of summarization systems, especially in Turkic languages with high expression diversity. Therefore, this study aims to create a dataset for video summarization in Turkish and to present a framework for generating gold summaries.

\section{Method}

\subsection{Dataset}

The video set consists of 82 lecture videos in Turkish on the topic of ``Data Structures and Algorithms.'' The videos were obtained from YouTube with the permission of the channel owners. All videos were included in the evaluation.

A total of 138 participants voluntarily watched these lectures and independently summarized the videos. The participants were computer science students aged 18--22. All participants were given precise instructions, with no restrictions on the length of their summaries, allowing them to include all points they considered important. After watching the videos, participants entered their summaries into online forms. After collecting the video summaries in this manner, summaries shorter than three sentences were removed from the dataset. At least 36 summaries were collected for each video, resulting in a total of 3,281 summaries.

\subsection{Automatic Meaning Unit Pyramid: AutoMUP}

This section explains how AutoMUP summaries are derived from human summaries. AutoMUP summaries reflect different levels of consensus-based content reliability derived from multiple human summaries. Among the three AutoMUP summaries generated for each video, only the highest-consensus summary (AutoMUP-1) is considered the gold summary in this study. Lower-consensus summaries (AutoMUP-2 and AutoMUP-3) are intentionally constructed from less frequently supported content units and are used to analyze the effect of consensus density on summary quality rather than serving as gold references.

\paragraph{Extraction and Embedding of Meaning Units.}
To extract informational content from the human summaries generated by participants, the texts were first divided into semantic units. In this process, the texts were automatically segmented at the sentence level using punctuation and line breaks, and units below a minimum length threshold were discarded. This procedure is fully automatic and does not involve any manual annotation or post-editing. Thus, each summary was transformed into singular and semantically coherent units.

The resulting semantic units were converted into dense embeddings using \texttt{paraphrase-multilingual-MiniLM-L12-v2}, a multilingual Sentence-Transformer model for Turkish. The embedding vector is calculated as follows:
\begin{equation}
\mathbf{e}_i = f(\mathbf{u}_i)
\end{equation}
Here, $\mathbf{u}_i$ represents a semantic unit, and $f(\cdot)$ denotes the embedding function. For each unit, the video ID, summary number, textual content, and embedding vector were recorded. This step ensures that human summaries are brought into a comparable form at both the linguistic and semantic levels.

\paragraph{Clustering-Based Consolidation of Meaning Units.}
Because different participants express the same content in various ways, instead of directly comparing the extracted meaning units, they are grouped according to their semantic similarities. For this purpose, hierarchical clustering based on cosine distance was applied to the embedded vectors. To determine the optimal threshold value in the clustering process, an automated threshold selection procedure was used; this procedure evaluates multiple distance thresholds and selects the value that yields a balanced cluster distribution.

As a result, content-like units were grouped together, and each cluster became a ``consensus unit of meaning'' for the relevant video. Two basic criteria were calculated for each cluster:
\begin{itemize}
    \item Support count: the number of different summaries contributing to the cluster,
    \item Support ratio: the ratio of this value to the total number of summaries.
\end{itemize}
This ratio was used as an empirical measure of significance, indicating the extent to which a unit of meaning was shared among participants.

The cluster center was calculated as follows:
\begin{equation}
\mathbf{c}_k = \frac{1}{|C_k|} \sum_{i \in C_k} \mathbf{e}_i
\end{equation}
Additionally, the expression with the embedding closest to the cluster center is designated as the representative unit of the cluster:
\begin{equation}
\mathbf{r}_k = \arg\min_{i \in C_k} \lVert \mathbf{e}_i - \mathbf{c}_k \rVert
\end{equation}

\paragraph{Consensus-Weighted Summary Construction.}
After semantically similar meaning units are clustered from multiple human summaries of the same video, a set of clusters is obtained, each representing a shared unit of meaning across participants. For a given video, the frequency of a cluster is quantified by counting the number of distinct human summaries that contribute at least one unit to that cluster. Clusters are ranked by decreasing support ratio, with ties broken by cluster size.

Based on this ordered ranking, three disjoint AutoMUP summaries are constructed for each video. Each summary is formed by selecting representative units from clusters according to their rank in the consensus hierarchy:
\begin{equation}
\text{Summary}^{(M)} = \{\mathbf{r}_{k_1}, \mathbf{r}_{k_2}, \ldots, \mathbf{r}_{k_M}\}
\end{equation}
Here, $M$ denotes the number of representative content units included in a summary. AutoMUP-1 consists of the top $M$ ranked clusters, AutoMUP-2 consists of the next $M$, and AutoMUP-3 consists of the following $M$. In all experiments, $M$ is fixed to 5, resulting in summaries of equal length. As clusters are ordered by decreasing consensus, the expected level of agreement decreases monotonically from AutoMUP-1 to AutoMUP-3. In this study, only AutoMUP-1 is treated as the gold summary, while the lower-consensus summaries are used to analyze the effect of consensus density on summary quality. This consensus-based selection follows the spirit of Pyramid-style frequency weighting, while being implemented in a fully automated, embedding-based framework.

\subsection{Comparison with LLM Summaries and Ablation Study}

The gold summaries generated by AutoMUP (AutoMUP-1) were compared with summaries produced by two strong LLMs (Flash~2.5 and GPT-5.1). Lower-consensus AutoMUP variants were used for controlled comparison, while two ablation settings were defined to analyze the contribution of individual components.

\section{Results}

\subsection{TR-EduVSum: Turkish Educational Video Summarization Dataset}

Video durations ranged from approximately 48 minutes to 3 minutes. The average video duration was calculated as 19 minutes and 18 seconds. Transcripts of the videos were extracted using the YouTube Subtitle Download tool and edited using the \texttt{fullstop-punctuation-multilingual-base} model. Each video contained a minimum of 378 and a maximum of 4,589 words. The average sentence length was calculated as approximately 15 words. Participants generated a minimum of 36 and a maximum of 53 independent summaries for each video (Table~\ref{tab:dataset_stats}). A total of 3,281 human summaries were included.

\begin{table}[t]
\centering
\small
\caption{Descriptive statistics of the TR-EduVSum dataset.}
\label{tab:dataset_stats}
\begin{tabular}{ll}
\hline
\textbf{Statistic} & \textbf{Value} \\
\hline
Total video duration & 26 h 23 min 24 sec \\
Average video duration & 19 min 18 sec \\
Median video duration & 15 min 24 sec \\
Minimum video duration & 3 min 06 sec \\
Maximum video duration & 48 min 12 sec \\
Total transcript word count & 161,464 \\
Average transcript length & 1,969 words \\
Median transcript length & 1,667 words \\
Minimum / Maximum words & 378 / 4,589 \\
Average word length & 5.63 characters \\
Average sentence length & 15.66 words \\
Summaries per video & 36--53 \\
Summary type & Abstractive \\
\hline
\end{tabular}
\end{table}

\subsection{Semantic Variability in Human Summaries}

\begin{figure}[t]
\centering
\includegraphics[width=\linewidth]{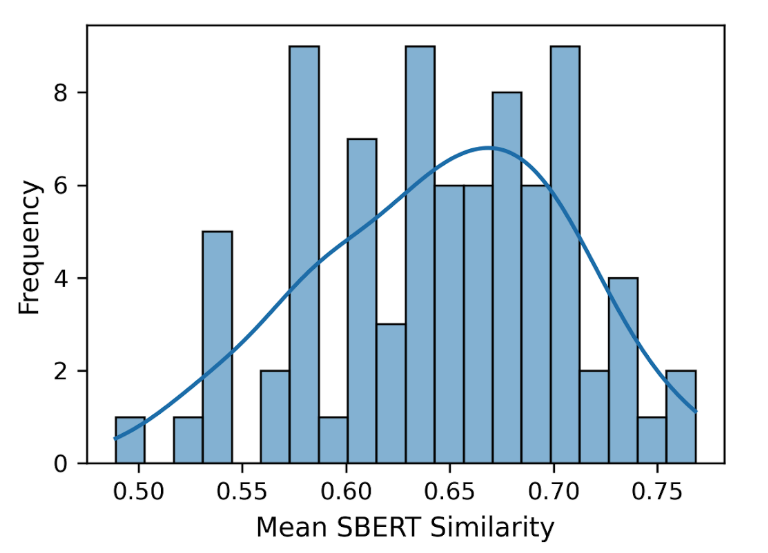}
\caption{Distribution of mean SBERT similarity across human summaries.}
\label{fig:sbert_distribution}
\end{figure}

The high expressive diversity of languages allows for the summarization of video content in many different ways. To reveal the variance between human summaries, the collected summaries were analyzed for semantic diversity. The similarity between the summaries produced by the participants for each video was calculated using SBERT (\texttt{paraphrase-multilingual-MiniLM-L12-v2}). Figure~\ref{fig:sbert_distribution} shows the distribution of the average pairwise SBERT similarity values between human summaries for each video.

Across the 82 videos, the average SBERT similarity value per video is approximately 0.65. The values vary approximately between 0.49 and 0.77, and the standard deviation of similarities between summary pairs lies in the range of 0.15--0.19 for most videos. According to these results, the similarity values between human summaries vary. This situation reveals that a single human summary cannot reliably represent the content. The applied consensus approach enables the identification of points commonly emphasized in the human summaries.

\subsection{Alignment between AutoMUP and LLM-based Summaries}

AutoMUP summaries were compared with summaries generated by Flash~2.5 and GPT-5.1 using BERTScore-F1 \citep{zhang2019}, ROUGE-L \citep{lin2004}, and BLEURT \citep{sellam2020}, along with embedded similarity metrics such as SBERT \citep{reimers2019}, SimCSE \citep{gao2021}, and Universal Sentence Encoder (USE) \citep{cer2018}. Table~\ref{tab:llm_alignment} shows that the AutoMUP-1 summary has the highest level of consensus and the highest average scores across all metrics for both LLMs. For example, the BERTScore-F1 values for Flash~2.5 were 0.872, 0.860, and 0.849 for AutoMUP-1, AutoMUP-2, and AutoMUP-3, respectively; while for GPT-5.1, these values were 0.865, 0.858, and 0.854. Similarly, a consistent decrease was observed from AutoMUP-1 to AutoMUP-3 in the SBERT, USE, ROUGE-L, and BLEURT scores. SimCSE scores exhibit a high level of agreement but limited variability, which may be attributed to a ceiling effect when comparing summaries derived from the same source content and to SimCSE’s lower sensitivity to differences in summary scope. Overall, the consistent downward trend across metrics indicates that ranking content units by consensus in AutoMUP produces quality-graded summaries and quantitatively validates the proposed framework.

\begin{table}[t]
\centering
\scriptsize
\setlength{\tabcolsep}{4pt}
\renewcommand{\arraystretch}{1.1}
\caption{Similarity between AutoMUP summaries and LLM-generated summaries. Higher scores indicate greater similarity.}
\label{tab:llm_alignment}
\begin{tabular}{lccc}
\hline
\textbf{Metric} & \textbf{A1} & \textbf{A2} & \textbf{A3} \\
\hline
\multicolumn{4}{l}{\emph{Flash 2.5}} \\
BERTScore-F1 & 0.872 & 0.860 & 0.849 \\
SBERT & 0.720 & 0.634 & 0.614 \\
SimCSE & 0.975 & 0.973 & 0.969 \\
USE & 0.711 & 0.660 & 0.630 \\
ROUGE-L & 0.246 & 0.166 & 0.144 \\
BLEURT & 0.405 & 0.309 & 0.257 \\
\multicolumn{4}{l}{\emph{GPT-5.1}} \\
BERTScore-F1 & 0.865 & 0.858 & 0.854 \\
SBERT & 0.655 & 0.585 & 0.585 \\
SimCSE & 0.968 & 0.968 & 0.967 \\
USE & 0.651 & 0.600 & 0.578 \\
ROUGE-L & 0.182 & 0.142 & 0.133 \\
BLEURT & 0.383 & 0.290 & 0.259 \\
\hline
\end{tabular}
\end{table}

\subsection{Effects of Consensus Weighting and Clustering on Summary Quality}

\begin{figure}[t]
\centering
\includegraphics[width=\linewidth]{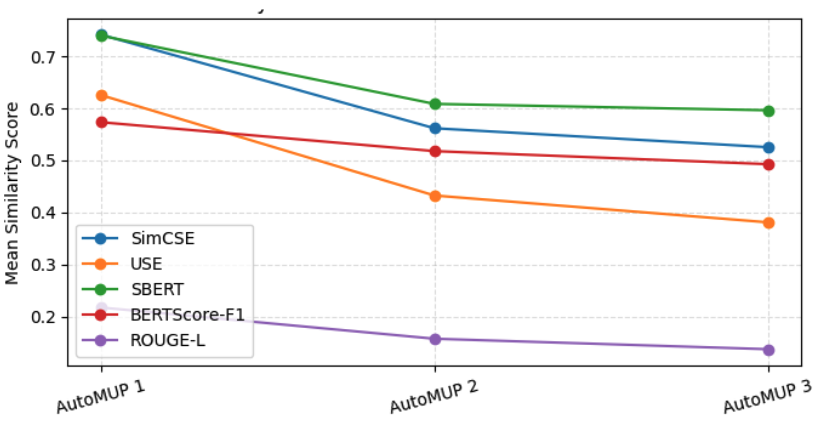}
\caption{Semantic similarity between AutoMUP summaries and human summaries across different consensus levels and ablation settings.}
\label{fig:ablation_similarity}
\end{figure}

AutoMUP summaries were compared with human summaries using SimCSE, USE, SBERT, BERTScore-F1, and ROUGE-L metrics. As expected, the gold summary produced by AutoMUP-1 achieves the highest alignment with human summaries across all metrics. The summary with the highest consensus (AutoMUP-1) showed the highest agreement with human summaries across all metrics. As the consensus rate of the summaries decreased, their similarity to human summaries also decreased (Figure~\ref{fig:ablation_similarity}). To investigate which components are decisive in determining the compatibility of AutoMUP-generated gold summaries with human summaries, two ablation conditions were defined: (i) No-Consensus (removal of consensus weighting) and (ii) No-Clustering (removal of the clustering step). The summaries generated under these ablation conditions were evaluated against human summaries using SimCSE, USE, SBERT, BERTScore-F1, and ROUGE-L metrics; the results are summarized in Table~\ref{tab:ablation}. 
AutoMUP-1 achieved the highest distributional semantic alignment with human summaries, with SimCSE (0.742), USE (0.626), SBERT (0.740), BERTScore-F1 (0.574), and ROUGE-L (0.217) values. This finding indicates that the generation of the same meaning unit by multiple participants creates a strong content signal, and that consensus weighting significantly increases the ability of gold summaries to represent human summaries. A consistent decrease in performance was observed across all metrics under the No-Consensus condition. SimCSE, USE, and SBERT values decreased to 0.488, 0.353, and 0.488, respectively. The fact that BERTScore-F1 (0.478) and ROUGE-L (0.116) scores are also below those of AutoMUP-1 indicates that, when consensus weighting is removed, the selected semantic units deviate from the content commonly emphasized in human summaries. These results reveal that the consensus weighting mechanism is a key component in preserving content representativeness and reflecting human agreement.

\begin{table*}[t!]
\centering
\small
\setlength{\tabcolsep}{6pt}
\renewcommand{\arraystretch}{1.15}
\caption{Comparison of AutoMUP and its ablation variants in terms of their alignment with human summaries (mean $\pm$ std). Higher scores indicate stronger alignment with human summaries. AutoMUP-1 corresponds to the gold summary in the proposed framework; other variants are included for ablation and comparison purposes.}
\label{tab:ablation}
\begin{tabular}{lccccc}
\hline
\textbf{Metric} & \textbf{AutoMUP-1} & \textbf{AutoMUP-2} & \textbf{AutoMUP-3} & \textbf{No-Clustering} & \textbf{No-Consensus} \\
\hline
SimCSE & 0.742 $\pm$ 0.049 & 0.562 $\pm$ 0.055 & 0.526 $\pm$ 0.086 & 0.625 $\pm$ 0.126 & 0.488 $\pm$ 0.104 \\
USE & 0.626 $\pm$ 0.068 & 0.433 $\pm$ 0.079 & 0.381 $\pm$ 0.093 & 0.521 $\pm$ 0.122 & 0.353 $\pm$ 0.105 \\
SBERT & 0.740 $\pm$ 0.059 & 0.609 $\pm$ 0.068 & 0.596 $\pm$ 0.087 & 0.625 $\pm$ 0.126 & 0.488 $\pm$ 0.104 \\
BERTScore-F1 & 0.574 $\pm$ 0.028 & 0.518 $\pm$ 0.027 & 0.493 $\pm$ 0.037 & 0.538 $\pm$ 0.043 & 0.478 $\pm$ 0.037 \\
ROUGE-L & 0.217 $\pm$ 0.035 & 0.157 $\pm$ 0.022 & 0.137 $\pm$ 0.027 & 0.200 $\pm$ 0.047 & 0.116 $\pm$ 0.028 \\
\hline
\end{tabular}
\end{table*}

In the No-Clustering condition, a significant improvement in semantic similarity metrics (SimCSE: 0.625, USE: 0.521, SBERT: 0.625) was observed compared to the No-Consensus setting. This result shows that semantic unit selection based solely on support frequency can still produce strong distributional semantic alignment. However, the fact that BERTScore-F1 (0.538) and ROUGE-L (0.200) values remain below the AutoMUP-1 level reveals that the clustering step plays a significant role in selecting more representative and expression-level consistent units by balancing content repetitions. The results clearly distinguish the functions of the two main components of AutoMUP. Consensus weighting provides the primary signal that ensures the gold summary reflects content commonly emphasized in human summaries, while clustering increases representativeness, reduces redundancy, and produces a more consistent surface structure. Together, these components enable AutoMUP to generate gold summaries that achieve the highest alignment with human summaries across semantic, surface-level, and content-based evaluation metrics.

\section{Conclusion}

This study presents the TR-EduVSum dataset, comprising 82 Turkish educational videos, multiple human-generated summaries, and model-generated summaries of these videos. The dataset contains a minimum of 36 and a maximum of 53 human summaries for each video. Two powerful LLMs with vision capabilities also summarized the videos, and these summaries were added to the dataset. A framework was then developed to generate gold summaries from the human summaries. Using the AutoMUP method, video summaries were generated from the human summaries at three graded quality levels, employing a weighted cluster ranking. In this study, only the AutoMUP-1 summary with the highest consensus level is considered the gold summary; AutoMUP-2 and AutoMUP-3 are used as comparative variants to analyze the method. The results show that while variance exists among the human summaries, the consensus-weighted structure of AutoMUP successfully reveals the common knowledge core of these summaries. AutoMUP-1 summaries achieved a high level of semantic agreement with summaries generated by two powerful LLMs with vision capabilities using the videos. Ablation analyses revealed that consensus weights form the basis for content selection, while clustering acts as a complementary component that increases representational power and consistency. In this study, a unique dataset comprising multiple human summaries was created to meet the need for a dataset for Turkish educational video summarization. A new framework for generating gold summaries (AutoMUP) was created from multiple human summaries, and it was shown that the gold summaries (AutoMUP-1) generated with this framework show high semantic similarity to strong LLM summaries and also overlap with human summaries. As a result, a Turkish educational video summarization dataset has been developed, containing multiple human summaries and gold summaries that can be used in Turkish video summarization studies. Because AutoMUP clusters content units in the SBERT space, the SBERT-based similarity score was reported only as a supporting measure during the evaluation phase; the overall performance of the method was interpreted through independent metrics such as SimCSE, USE, and BERTScore. AutoMUP summaries consistently exhibited high semantic overlap with LLM summaries across all independent metrics. SBERT results also support this trend. The videos included in the study are limited to lecture videos in the field of ``Data Structures and Algorithms.'' Differences in video lengths and the number of videos published by instructors for lectures may have affected the quality of summaries produced by participants. The framework created is based on consensus among human summaries. This design may overlook minority but relevant viewpoints; however, this trade-off was made to capture consistent and repeatable content.

\section*{Limitations}

Because AutoMUP clusters content units in the SBERT space, the SBERT-based similarity score was reported only as a supporting measure during the evaluation phase; the overall performance of the method was interpreted through independent metrics such as SimCSE, USE, and BERTScore. AutoMUP summaries consistently exhibited high semantic overlap with LLM summaries across all independent metrics. SBERT results also support this trend. The videos included in the study are limited to lecture videos in the field of "Data Structures and Algorithms." Differences in video lengths and the number of videos published by instructors for lectures may have affected the quality of summaries produced by participants. The framework created is based on consensus among human summaries. This is a situation that can affect accuracy in some cases, which is often misunderstood. This design may overlook minority but relevant viewpoints. However, this has been sacrificed to capture consistent and repeatable content.

\section*{Ethics Statement}

This work adheres to the ACL Ethics Policy and follows established standards for responsible research in natural language processing. All videos used in the dataset are publicly available educational materials on YouTube. Human summaries used in this study were collected voluntarily from annotators who were informed about the purpose of the research. No demographic information was recorded, and no sensitive user data was processed. All annotators were free to withdraw at any point. The study complies with standard data protection and privacy guidelines. The research protocol, including the collection and use of human-written summaries, was reviewed and approved by the relevant institutional authorities, and all necessary permissions were obtained prior to data collection. Finally, this work aims to support fair, reproducible, and transparent evaluation practices in multilingual summarization research, particularly in low-resource settings.

\section*{ Acknowledgement}

This paper is derived from ongoing doctoral research conducted at İzmir Kâtip Çelebi University.

\bibliography{custom}
\bibliographystyle{acl_natbib}

\end{document}